\documentclass[conference]{IEEEtran}
\IEEEoverridecommandlockouts
\usepackage{cite}
\usepackage{amsmath,amssymb,amsfonts}
\usepackage{algorithmic}
\usepackage{graphicx}
\usepackage{textcomp}
\usepackage{xcolor}
\usepackage{multirow}
\usepackage{makecell}
\usepackage{tabularx}
\usepackage[colorlinks,linkcolor=blue,anchorcolor=blue,citecolor=blue]{hyperref}
\usepackage{amsfonts}
\usepackage{flushend}
\begin{document}
\renewcommand{\figurename}{Figure}
\title{Learning Friction Model for Tethered Capsule Robot\\
%\thanks{\textsuperscript{*}Corresponding author\\
 %Email address: limiao712@gmail.com}
}

\author{\IEEEauthorblockN{Yi Wang}
\IEEEauthorblockA{\textit{School of Power and Mechanical} \\
\textit{Engineering}\\
\textit{Wuhan University}\\
Wuhan, China \\
yi.wang@whu.edu.cn}
\and
\IEEEauthorblockN{Yuchen He}
\IEEEauthorblockA{\textit{School of Mechanical} \\
\textit{Engineering and Automation}\\
\textit{Wuhan Textile University}\\
Wuhan, China \\
heyuchen2021@outlook.com}
\and
\IEEEauthorblockN{Xutian Deng}
\IEEEauthorblockA{\textit{School of Power and Mechanical} \\
\textit{Engineering}\\
\textit{Wuhan University}\\
Wuhan, China \\
dengxutian@whu.edu.cn}
\and
\IEEEauthorblockN{Ziwei Lei}
\IEEEauthorblockA{\textit{School of Power and Mechanical} \\
\textit{Engineering}\\
\textit{Wuhan University}\\
Wuhan, China \\
leiziwei@whu.edu.cn}
\and
\IEEEauthorblockN{Yiting Chen}
\IEEEauthorblockA{\textit{School of Power and Mechanical} \\
\textit{Engineering}\\
\textit{Wuhan University}\\
Wuhan, China \\
chenyiting@whu.edu.cn}
\and
\IEEEauthorblockN{Miao Li\textsuperscript{*}}
\IEEEauthorblockA{\textit{School of Power and Mechanical} \\
\textit{Engineering}\\
\textit{Wuhan University}\\
Wuhan, China \\
limiao712@gmail.com}
}
\maketitle

\begin{abstract}
With the potential applications of capsule robots in medical endoscopy, accurate dynamic control of the capsule robot is becoming more and more important. In the scale of a capsule robot, the friction between capsule and the environment plays an essential role in the dynamic model, which is usually difficult to model beforehand. In the paper, a tethered capsule robot system driven by a robot manipulator is built, where a strong magnetic Halbach array is mounted on the robot's end-effector to adjust the state of the capsule. To increase the control accuracy, the friction between capsule and the environment is learned with demonstrated trajectories. With the learned friction model, experimental results demonstrate an improvement of 5.6\% in terms of tracking error.  \\

\emph{Keywords}--- tethered capsule robot, magnetic actuation, friction model
\end{abstract}

%\begin{IEEEkeywords}
%capsule robot, magnetic actuation, friction model
%\end{IEEEkeywords}

\section{introduction}

Capsule endoscopy has been extensively studied since its first invention by Iddan \cite{ref1}. The actuation for capsule robots can be generally divided into two different categories: active and passive. The actuator of the active capsule robot is delicately embedded into its size-limited structure, which results in its vulnerability and complicated manufacturing process. Compared with the active capsule robot, the passive capsule robot has a simpler internal structure and is more stable for practical applications.

%There are three popular methods to provide external field strength. The first method is to use the Helmholtz coil. Salehizadeh \emph{et al.} used coil microrobot to move in any trajectory \cite{ref2}. Yongshun Zhang \emph{et al.} used coil to control the rotation and movement of spherical capsule robot \cite{ref3}. Cheong Lee \emph{et al.} used coil to control the movement of capsule robot in the intestinal tract \cite{ref4}. The second is to use a combination of external permanent magnet and electromagnetic coil. Taddese \emph{et al.} used a permanent magnet and coil combination to control the capsule robot steering and moving in PVC pipe \cite{ref5}\cite{ref6}. The third is to use an external permanent magnet. Mahoney \emph{et al.} used an external permanent magnet to control the three degrees of freedom and two directions of the capsule robot \cite{ref7}\cite{ref8}. The external permanent magnet drive has the advantages of no energy consumption, constant magnetic induction, and ease to control. There are three main external magnetic field control drives: the transplanting platform machine, the handheld drive, and the manipulator. Chen Fanji \emph{et al.} used the transplanting platform to control the movement of the coil \cite{ref9}. GI Shih lien \emph{et al.} used a hand-held driver to control the motion of the capsule robot \cite{ref10}. Taddese \emph{et al.} used a manipulator arm as a driver \cite{ref5}\cite{ref6}\cite{ref7}. Among them, the manipulator is easy and sensitive to control and has a variety of gestures.

There are three popular methods to provide external field strength. The first method is to use the Helmholtz coil. A coiled microrobot is designed to move in all the directions in \cite{ref2}. The spherical capsule robot controlled by the coil can rotate and translate \cite{ref3}. A coil-controlled capsule robot is used in the intestinal movement \cite{ref4}. The second method is to use a combination of an external permanent magnet and an electromagnetic coil, which is used to control the steering and movement of capsule robot in PVC (polyvinyl chloride) pipe \cite{ref5}\cite{ref6}. The third method is to use an external permanent magnet. A capsule robot with three degrees of freedom and two directions is controlled base on this method \cite{ref7}\cite{ref8}. The external permanent magnet drive has the advantages of no energy consumption, constant magnetic induction, and ease to control. There are three main external magnetic field control drives: the transplanting platform machine \cite{ref9}, the handheld drive \cite{ref10}, and the manipulator \cite{ref5}\cite{ref6}\cite{ref7}. Among them, the manipulator is easy and sensitive to control and has a variety of gestures, which is adopted in this paper.

For the transmission of signals, capsule robots can be divided into two categories: wireless and tethered. For a wireless capsule robot, the image signal transmission has a certain delay because it relies on the internal image transmitting device and the external image receiving device \cite{ref1}. Doctors observe the reconstruction of images captured by wireless capsule robots. Tethered capsule robots have a stable video transmission channel and a lighting energy supply. Despite of the merits of tethered capsule robots, it is also difficult to control the whole system due to the friction and the drag force from the tethered. The mathematical relationship between the manipulator and the capsule robot established in \cite{ref7}, which corresponded the joint speed of the manipulator to the position of the capsule robot, and could control five degrees of freedom, including three directions of movement and two directions of rotation. In addition, Hall-effect sensors are added to the capsule robot to detect the real-time position and to introduce feedback direction control to realize the closed-loop control of the capsule robot \cite{ref4}\cite{ref11}\cite{ref12}. However, the environmental friction in the tethered capsule robot motion model is still not clearly taken into account.

\begin{figure}[htbp]
\centerline{\includegraphics[width=3in]{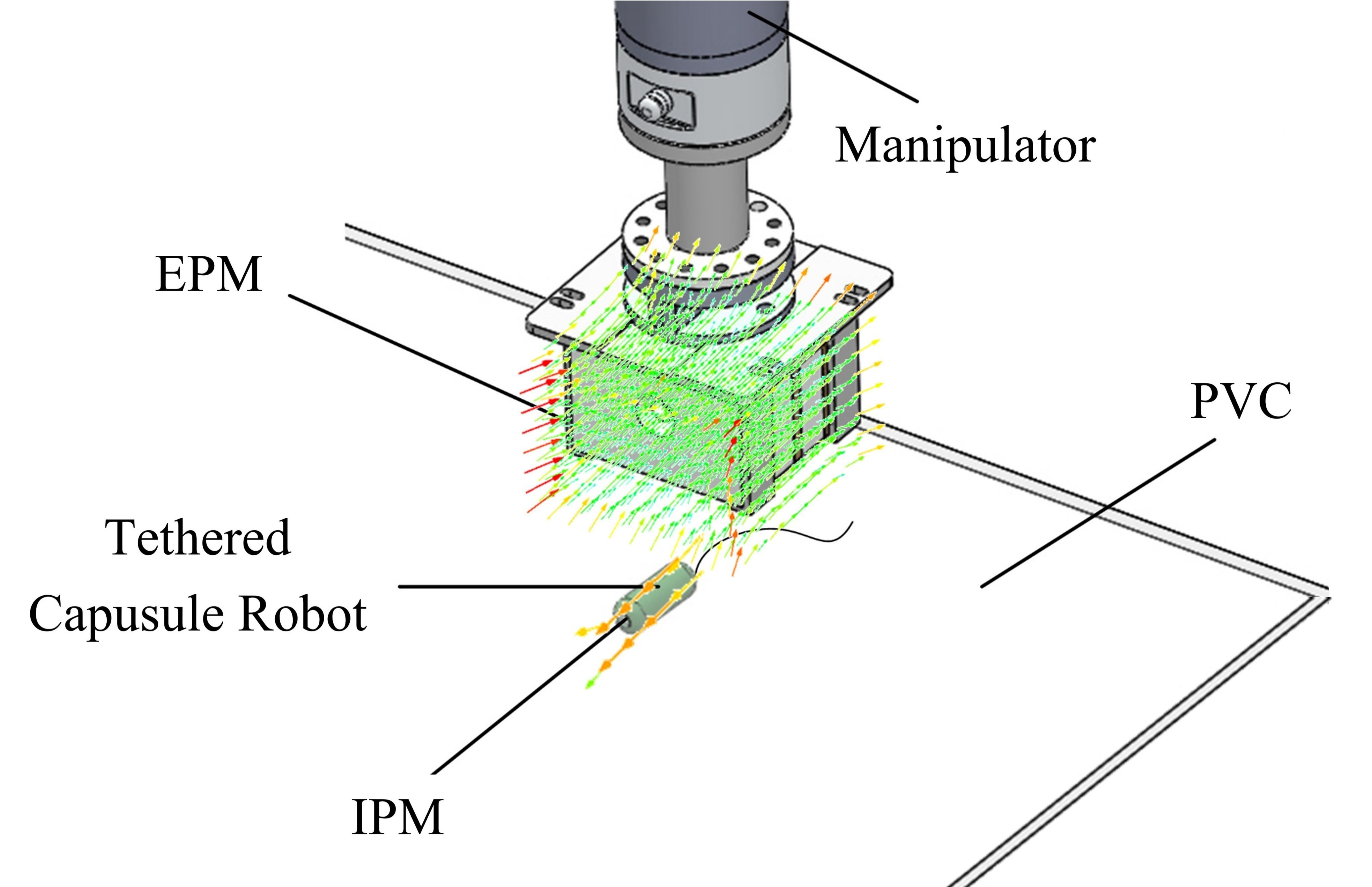}}
\caption{A scene diagram in the simulated environment: At the end of the manipulator is a external permanent magnet (EPM), and an inner permanent magnet (IPM) is inside the terthered capsule robot. PVC is an obstacle between IPM and EPM.   \textbf{} }
\label{fig1}
\end{figure}

In this paper, we propose a dynamic model with friction compensation for the planar motion of the capsule robot. Based on the theoretical control scheme of wireless capsule robot proposed by Mahoney \emph{et al.} \cite{ref7}. We extracted a dynamic model with friction compensation for the control of the tethered capsule robot. Furthermore, we build a magnetic-driven capsule robot system, which includes a six-axis manipulator, a capsule robot, a camera, and a PVC partition (Fig.~\ref{fig1}).

This paper is organized as follows: In Section \uppercase\expandafter{\romannumeral2}, the theory of the capsule friction model is proposed. In Section \uppercase\expandafter{\romannumeral3}, the hardware and software of the system are studied. In Section \uppercase\expandafter{\romannumeral4}, the trajectory tracking and magnetic field simulation of the capsule robot are studied and the specific expression of the friction theory in the current experimental environment through experiments is demonstrated. In Section \uppercase\expandafter{\romannumeral5}, the whole research work and results of this article are summarized. The main contribution of our research is to give a learning friction model, which can be applied to translation of tethered capsule robot between any two contact medium.In addition, a strong magnetic Halbach array is applied to improve the external field intensity for a capsule robot.

\section{Methods}

A friction dynamic model between the capsule robot and the contact interface is proposed to optimize the motion control of the capsule robot attached to different media. We expect to obtain a more specific friction relationship between capsule robot and environment to improve the control accuracy.

\subsection{Dynamics Model of Magnetron Capsule}\label{AA}
%``Tab.~\ref{tab1}'' outlines the nomenclature used throughout the paper. The (Figure~\ref{fig4}) shows the force diagram of magnetron capsule. 

According to the application of the point dipole model in \cite{ref7}\cite{ref11}, the theoretical dynamics model of magnetron capsule is defined as:
\begin{equation}
B(x)\ddot x + C(x,\dot x)\dot x + G(x) = {\tau _m}(x,q)\label{eq1}
\end{equation}

where $x \in {\mathbb{R}^3}$ is the capsule pose (position and orientation) and $q \in {\mathbb{R}^6}$ embeds the robot joint variables; matrices $B(x)$, $C(x,\dot x)$, $G(x)$ are the respective inertia, Coriolis matrix and gravity. The vector ${\tau _m}(x,q) \in {\mathbb{R}^6}$ represents the magnetic force and torque exerted by the EPM on the IPM.  Our aim is to find $q$ such that $x$ approaches a desired value ${x_d}$. 

The differential of \eqref{eq1} can be deduced as the following expression:
\begin{equation}
{\dot \tau _m} = \frac{{\partial \tau (x,q)}}{{\partial x}}\dot x + \frac{{\partial \tau (x,q)}}{{\partial q}}\dot q = {J_x}\dot x + {J_q}\dot q\label{eq2}
\end{equation}

${\tau _m}$ is the state variable of the control system\cite{ref11}. It is obtained by \eqref{eq1} and \eqref{eq2}:
\begin{equation}
\left\{ {\begin{array}{*{20}{c}}
{B(x)\ddot x + C(x,\dot x)\dot x + G(x) = \tau }\\
{\dot \tau  = {J_x}\dot x + {J_q}\dot q + \dot \kappa }
\end{array}} \right.\label{eq3}
\end{equation}

$\tau $ is the actual force and moment on the EPM. Other parameters are shown in Table.~\ref{tab1}.

The $\dot \kappa $ mentioned in \eqref{eq3} is put as equivalent to a straight connecting rod in the simulation environment \cite{ref11}, which is not verified in the real environment. In the real environment, $\dot \kappa $ is judged to represent the friction between the capsule and the environment, including kinetic friction force ${{\bf{f}}_{\bf{k}}}$ and static friction force (lead drag force ${{\bf{f}}_{\bf{s}}}$). 

\subsection{Planar dynamics model of capsule}\label{BB}
The theoretical kinetic model of magnetron capsules can be expressed as:
\begin{equation}
\begin{array}{l}
{\bf{F + }}{{\bf{f}}_{\bf{s}}}{\bf{ + }}{{\bf{f}}_{\bf{k}}}{\bf{ = }}m{\bf{a}},{\rm{   }}{{\bf{f}}_{\bf{k}}}{\bf{ = }}{{\bf{F}}_{\bf{N}}} \cdot \mu \\
\\
{\bf{F + }}{{\bf{f}}_{\bf{s}}}{\bf{ + }}{{\bf{F}}_{\bf{N}}} \cdot \mu {\bf{ = }}m{\bf{a}}
\end{array}\label{eq4}
\end{equation}

\begin{figure}[htbp]
\centerline{\includegraphics[width=3.5in]{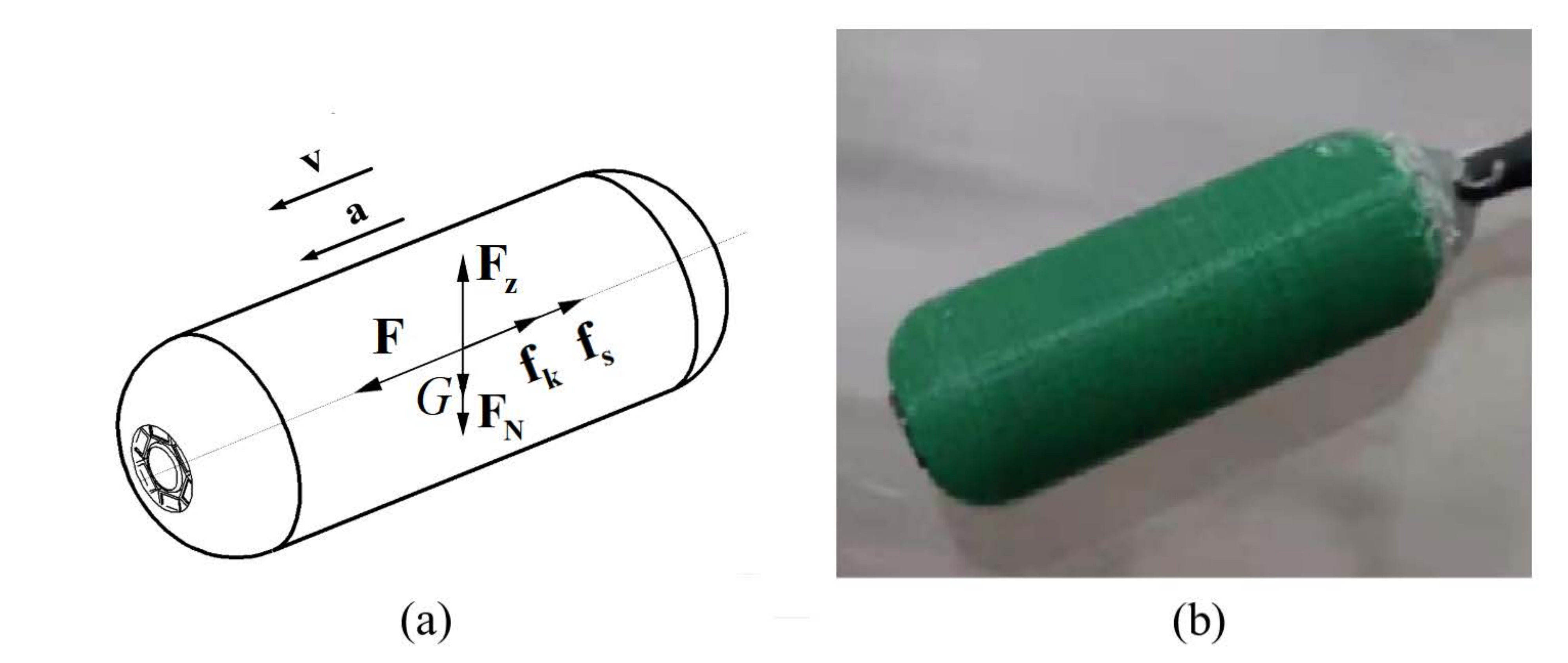}}
\caption{The force diagram of the magnetron capsule: (a) Schematic diagram of the force of the capsule robot in the working state. (b) The top view of the capsule robot's working state.}
\label{fig2}
\end{figure}

The force diagram of the model is shown in Fig.~\ref{fig2}(a), and the other parameters are shown in Table.~\ref{tab1}.

%双栏放表格
%\begin{table*}[htbp]
%\caption{Nomenclature Used}
%\begin{center}
%\begin{tabular}{|c|c|}
%\hline
%\textbf{Symbol}&{\textbf{Description}} \\
%%\cline{2-2} 
%%\textbf{Head} & \textbf{\textit{Table column subhead}}& \textbf{\textit{Subhead}}& \textbf{\textit{Subhead}} \\
%\hline
%%$x \in {R^3}$& Capsule Gesture (position and orientation)  \\
%%$q \in {R^6}$& Robot Joint Variables  \\
%%$B(x)$& Respective Inertia   \\
%%$C(x,\dot x)$& Coriolis Matrix  \\
%%$G(x)$& Gravity  \\
%%${\tau _m}(x,q)$& Magnetic Dipole Force and Torque (EPM on IPM)\\
%%$\tau $& Actual Force and Torque on the Capsule  \\
%$\kappa $& Friction between Lead and Environment  \\
%${\bf{F}}$& Traction Force of Capsule in Movement Direction(Theoretical Simulation Calculation)\\
%${{\bf{F}}_{\bf{z}}}$& Magnetic Force of capsule in Vertical Direction(Theoretical Simulation Calculation)\\
%${{\bf{F}}_{\bf{N}}}$& Magnetic Force of capsule in Vertical Direction(Theoretical calculation)\\
%${{\bf{f}}_{\bf{s}}}$& Static Friction Force($<$ 0.01N,not be considered)\\
%$m$& Capsule Mass (20.9g)\\
%${\bf{a}}$& Capsule Acceleration\\
%${\bf{v}}$& manipulator Speed (Preset Experimental Speed)\\
%${{\bf{v}}^{\bf{*}}}$& The Typical Speed,about 0.2m/s [15]\\
%${{\bf{D}}_{\bf{c}}}$& Critical Slip Length [15]\\
%$a$& System Constant [15,16]\\
%$b$& System Constant [16,16]\\
%${\mu _0}$& Sliding Friction Coefficient\\
%$c$& Friction System Relationship to be Solved\\
%\hline
%%\multicolumn{4}{l}{$^{\mathrm{a}}$Sample of a Table footnote.}
%\end{tabular}
%\label{tab1}
%\end{center}
%\end{table*}

%单栏放表格

\begin{table}[htbp]
\renewcommand\arraystretch{1.5}%设置行距
\caption{Nomenclature Used}
\begin{center} 

%\begin{tabular}{|c|c|}%有框
\begin{tabularx}{\linewidth}{ c c }%无框
\hline
\textbf{\space\space\space Symbol \space\space\space\space}&\textbf{{Description}} \\
%\cline{2-2} 
%\textbf{Head} & \textbf{\textit{Table column subhead}}& \textbf{\textit{Subhead}}& \textbf{\textit{Subhead}} \\
\hline
%$x \in {R^3}$& Capsule Gesture (position and orientation)  \\
%$q \in {R^6}$& Robot Joint Variables  \\
%$B(x)$& Respective Inertia   \\
%$C(x,\dot x)$& Coriolis Matrix  \\
%$G(x)$& Gravity  \\
%${\tau _m}(x,q)$& Magnetic Dipole Force and Torque (EPM on IPM)\\
%$\tau $& Actual Force and Torque on the Capsule  \\
$\kappa $& Friction between Lead and Environment  \\
%\hline
\multirow{2}{*}{}{${\bf{F}}$}& \makecell[c]{Traction Force of Capsule in Movement Direction}\\
%\hline
\multirow{2}{*}{}{${{\bf{F}}_{\bf{z}}}$}& \makecell[c]{Magnetic Force of capsule in Vertical Direction}\\
%\hline
\multirow{2}{*}{}{${{\bf{F}}_{\bf{N}}}$}&  \makecell[c]{Support Force of capsule by PVC}\\
%\hline
${{\bf{f}}_{\bf{s}}}$& Static Friction Force \\
%\hline
${{\bf{f}}_{\bf{k}}}$& Kinetic Friction Force\\
%\hline
$m$& Capsule Mass (20.9g)\\
%\hline
${\bf{a}}$& Capsule Acceleration\\
%\hline
${\bf{v}}$& Manipulator Speed (Preset Experimental Speed)\\
%\hline
${{\bf{v}}^{\bf{*}}}$& The Typical Speed, about 0.2m/s \cite{ref15}\\
%\hline
${\bf{D_{c}}}$& Critical Slip Length \cite{ref15}\\
%\hline
$a$& System Constant \cite{ref15}\cite{ref16}\\
%\hline
$b$& System Constant \cite{ref15}\cite{ref16}\\
%\hline
${\mu _0}$& Sliding Friction Coefficient\\
%\hline
$c$& Equation of Friction System to be Solved.\\
\hline
%\multicolumn{4}{l}{$^{\mathrm{a}}$Sample of a Table footnote.}
\end{tabularx}
\label{tab1}
\end{center}
\end{table}

In particular, kinetic friction and static friction have been proved to be relative. These two concepts can be replaced by a view of rate-dependent friction, the law of friction of rocks \cite{ref15}, which is generally applicable to materials including polymers, glasses, etc. In \cite{ref15}, the coefficient of friction is related to instantaneous velocity ${\bf{v}}$ and state variable $\theta $.
\begin{equation}
\mu  = {\mu _0} - a\ln \left( {\frac{{{{\bf{v}}^{\bf{*}}}}}{{{\bf{|v|}}}} + 1} \right) + b\ln \left( {\frac{{{{\bf{v}}^{\bf{*}}}{\rm{\theta }}}}{\bf{D_{c}}} + 1} \right)\label{eq5}
\end{equation}

For the state variable $\theta $ in law \eqref{eq5}, the following equation is available,
\begin{equation}
\dot{\theta}=1-\left(\frac{\bf{|v|} \theta}{\bf{D_{c}}}\right)\label{eq6}
\end{equation}

The constants a and b in law \eqref{eq5} are both positive numbers, with the order of magnitude from ${\bf{D_{c}}}$ , and the order of magnitude of ${\bf{D_{c}}}$ under the experimental conditions is 10um. In the static case, $\theta =t\ $ is true. Therefore, the state variable can be interpreted as the average time from the beginning of the movement of the micro-contact. When the moving speed is constant ${\bf{|v|}}$ and the initial condition is $\theta (0)={{\theta }_{0}} $, the solution of equation \eqref{eq6} is:

\begin{equation}
\theta(t)=\frac{\bf{D_{c}}}{\bf{|v|}}+\left(\theta_{0}-\frac{\bf{D_{c}}}{\bf{|v|}}\right) \exp \left(-\frac{{\bf{|v|}} t}{\bf{D_{c}}}\right)\label{eq7}
\end{equation}

All existing micro-contacts along the sliding length ${D_{c}}$ where ${D_{c}}$ is located are destroyed and replaced by new micro-contacts. After this transformation process,\begin{equation}
\theta(\infty)=\frac{\bf{D_{c}}}{\bf{v}}\label{eq8}
\end{equation}

This is true, which is consistent with the interpretation of state variable 0 as an age variable. In this case, the steady-state value of $\theta $ is equal to the average contact time of micro-contact. For steady-state sliding, the friction coefficient is the formula \eqref{eq9}:
\begin{equation}
\mu=\mu_{0}- c\ln \left(\frac{{\bf{v}}^{*}}{\bf{|v|}}+1\right)\label{eq9}
\end{equation}

The Dieterich-Ruina friction law provides a good description of the unsteady state transition process. We consider a friction process with a sliding speed of v. According to \eqref{eq7}, the steady-state friction coefficient is equal to \eqref{eq8}, and c expresses the unsteady-state friction system equation to be solved.

$\left| {\bf{v}} \right| <  < {{\bf{v}}^*}$, the expression of friction law \eqref{eq10} can be rewritten as:
\begin{equation}
\mu  = {\mu _0} - c\ln \left( {\frac{{{{\bf{v}}^{\bf{*}}}}}{{{\bf{|v|}}}} } \right)\label{eq10}
\end{equation}

\begin{equation}
{\bf{F + }}{{\bf{f}}_{\bf{s}}}{\bf{ + }}{{\bf{F}}_{\bf{{\bf N}}}} \cdot \left[ {{\mu _0} - c \cdot \ln \left( {\frac{{{{\bf{v}}^{\bf{*}}}}}{{{\bf{|v|}}}} } \right)} \right] = m{\bf{a}}\label{eq11}
\end{equation}

The drag force ${{\bf{f}}_{\bf{s}}}$ of the system is obtained using the tension dynamometer to measure, which is 0.03N. ${\bf{F}}$ and ${{\bf{F}}_{\bf{z}}}$ are obtained by theoretical simulation. From \eqref{eq12}, the ambient friction force of the capsule robot can be compensated by adjusting the ${\bf{v}}$ of the manipulator, to determine the force of the capsule robot, furthermore, the instantaneous acceleration of the capsule robot can be controlled, but this friction relation $c$ must be determined in advance, which is obtained through physical experiments as shown in the following section.

\section{Experimental system setup}

\begin{figure}[htbp]
\centerline{\includegraphics[width=3.5in]{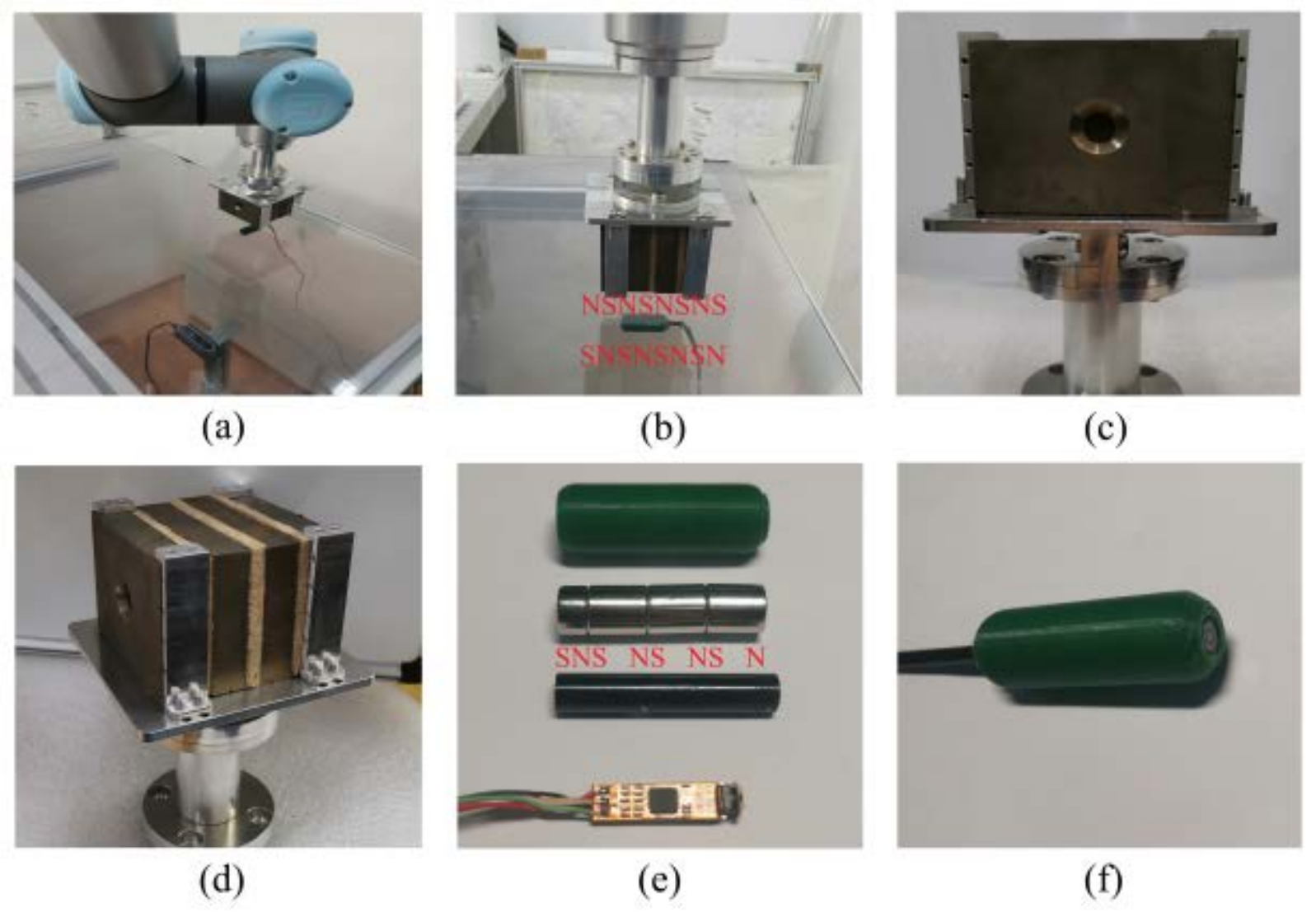}}
\caption{Structure of system: (a) Schematic diagram of the overall scene. (b) Working diagram of EPM and IPM. (c) Front view of external permanent magnet. (d) Axonometric view of external permanent magnet. (e) A detail drawing of the capsule robot. (f) Assembled capsule robot.}
\label{fig3}
\end{figure}

%\subsection{The construction of hardware system}\label{AA}
It can be seen the overall hardware system from (Fig.~\ref{fig3}(a)(b)). The external permanent magnet (EPM) terminal designed in this paper is composed of four strong magnets in a Halbach Array \cite{ref13} (As shown in Fig.~\ref{fig3}(c)(d)), which is fixed on the end of the manipulator through 6061 aluminum alloy. The strong magnetic terminal can provide a strong and stable magnetic field. The EPM is mounted at the end of a robot arm with 6 DOFs (degrees of freedom). The capsule robot is composed of a camera module, circuit board sealed shell, inner permanent magnet (IPM), and PLA (polylactic acid)
shell. IPM design inspiration comes from Mahoney \emph{et al.}( A solid cylindrical magnet as IPM in \cite{ref7}), and F. Zhang \emph{et al.} (Two disk magnets as IPM in \cite{ref14}).  The weight, length, and diameter of the capsule robot are 20.9g, 36mm, and 14mm respectively. The specific structure is shown in Fig.~\ref{fig3}(e)(f). During the movement, the capsule robot is attached to the PVC plate by external magnetic force. To explore the dynamic model with friction compensation of the capsule robot, we use the planar friction between a PVC plate and the capsule to provide a fixed kinetic friction coefficient. A transparent PVC plate is selected to facilitate camera calibration and collect the motion information of the capsule robot.

\begin{figure}[htbp]
\centerline{\includegraphics[width=3.5in]{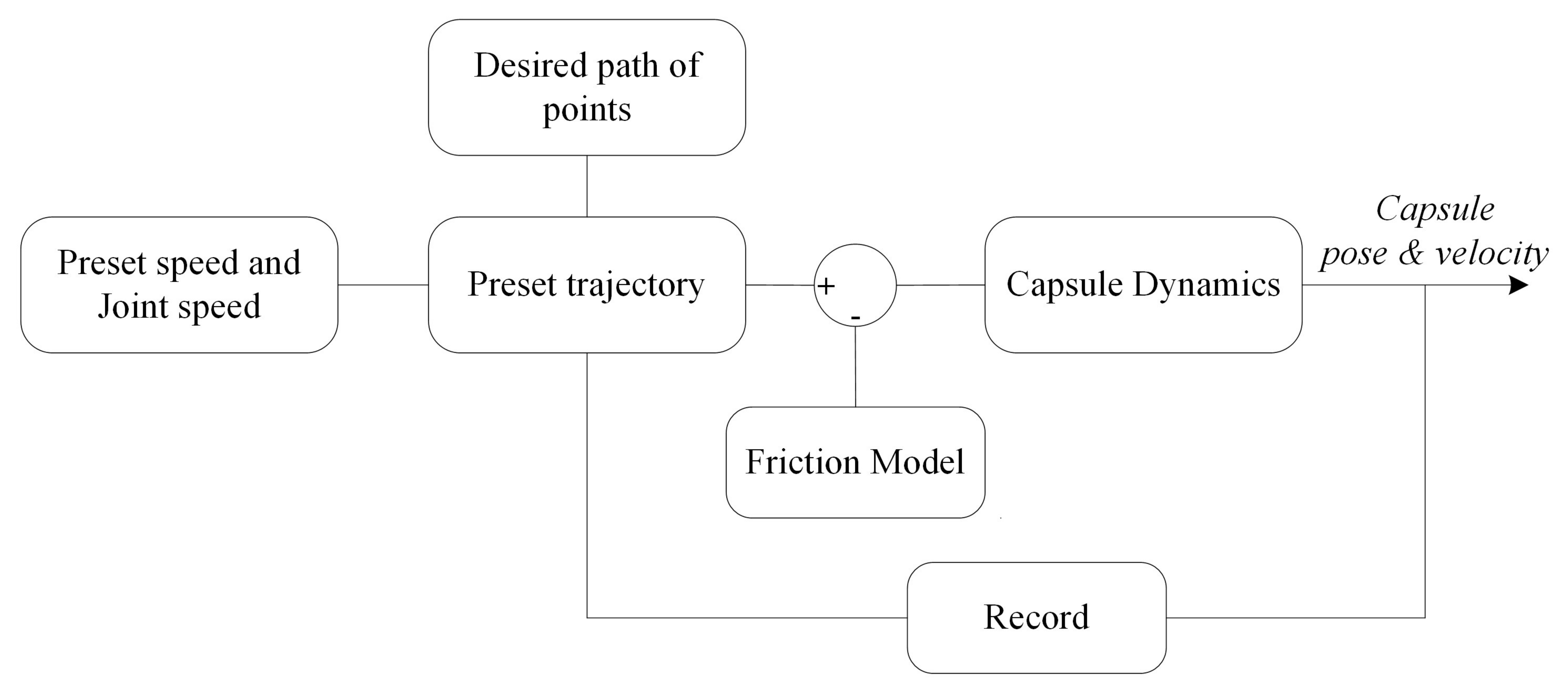}}
\caption{ Flowchart of magnetron motion.}
\label{fig4}
\end{figure}

The control diagram of the whole system is shown in Fig.~\ref{fig4}. The force of the capsule robot is different when the relative positions of the IPM and EPM are different. To control capsule robot to motion under trajectory map, we need to control the manipulator (EPM) motion.first the preset speed and joint speed are used, second, the trajectory is broken up into many points. The manipulator carries on from one point to the next until the last point. The friction between the capsule robot and the environment can be compensated by adjusting the speed of the manipulator through the friction model. The position of IPM and EPM can be saved in real time.

%一栏
%%两栏
%\begin{figure*}[htbp]
%\centerline{\includegraphics[width=5in]{3a}}
%\caption{Magnetic capsule control system.}
%\label{fig3}
%\end{figure*}

\section{Experiments}
\subsection{Tracking of Capsule Trajectory}
\begin{figure}[htbp]
\centerline{\includegraphics[width=3.5in]{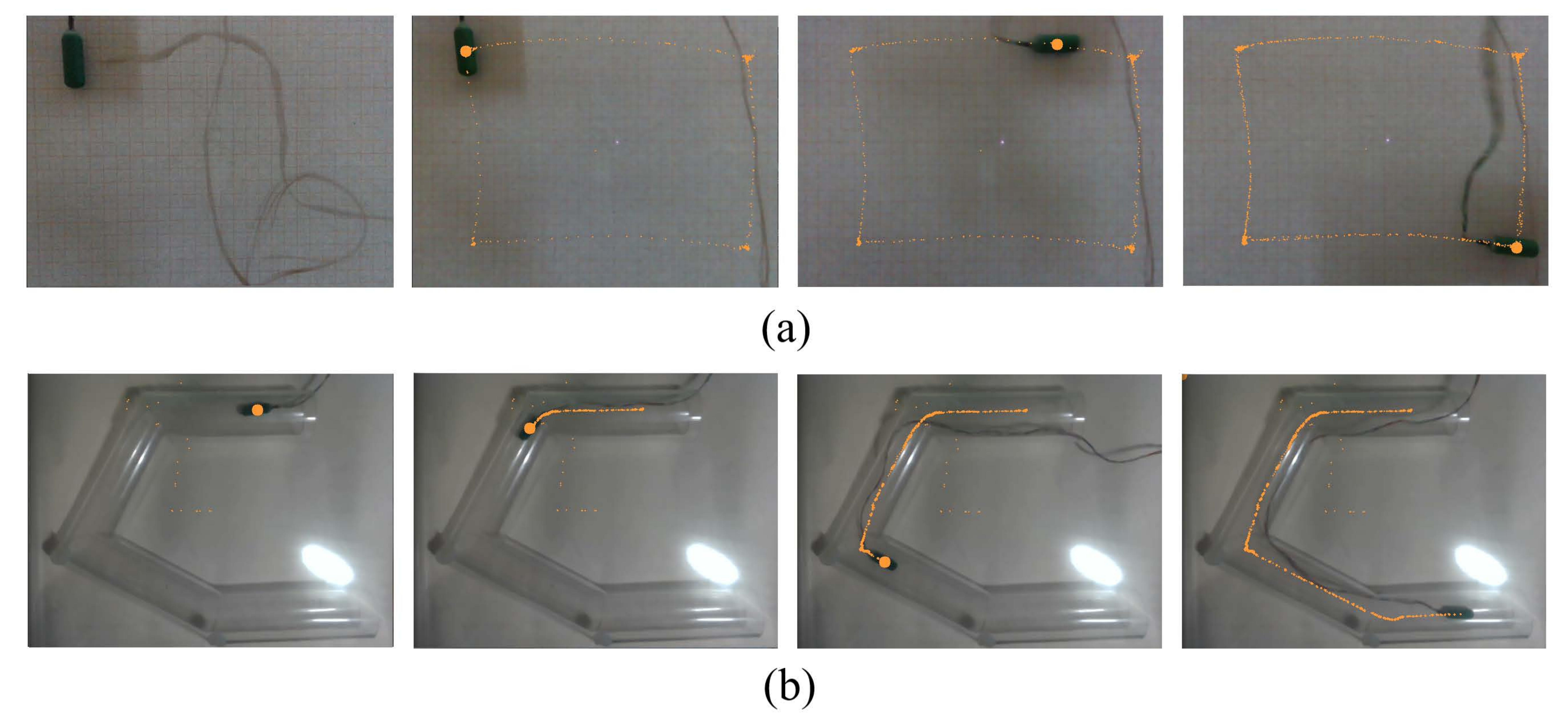}}
\caption{The trajectory tracking process of the capsule robot: (a) The rectangular trajectory of the capsule robot. (b) The pipeline trajectory of the capsule robot.}
\label{fig5a}
\end{figure}

The coordinates of the end of the manipulator and the capsule robot are read at the same frequency, which ensures that the position information is saved based on the same time series. Fig.~\ref{fig5a} show the capsule tracking process of the capsule robot on the tablet and inside the pipeline.

% The RGB information of each frame of the camera is obtained in real-time to obtain the trajectory points of the capsule robot in the process of movement.

\subsection{Finite Element Simulation: Obtaining ${\bf{F}}$ and ${{\bf{F}}_{\bf{z}}}$}

The finite element analysis software Maxwell is based on Maxwell's equations. The electromagnetic force between magnets is calculated by using the equivalent volume current method for permanent magnets and the change of virtual displacement magnetic energy. In this control experiment, the forces {\bf{F}} and ${{\bf{F}}_{\bf{z}}}$ of IPM can be obtained. The parameters of EPM and IPM within are shown in Table.~\ref{tab2}.

\begin{table}[htbp]
\renewcommand\arraystretch{1.5}%设置行距
\caption{Parameters of EPM and IPM}
\begin{center}
\begin{tabularx}{\linewidth}{c c c }
\hline
%\textbf{Parameter}&\multicolumn{3}{|c|}{\textbf{Table Column Head}} \\
\textbf{Parameter}&\textbf{\space\space\space\space\space\space EPM \space\space\space\space\space\space}&\textbf{\space\space\space\space\space\space IPM\space\space\space\space\space\space }\\
%\cline{2-4} 
%\textbf{Head} & \textbf{\textit{Table column subhead}}& \textbf{\textit{Subhead}}& \textbf{\textit{Subhead}} \\
\hline
\text{Magnetization (T)}& 1.26 & 1.26 \\
\text{Length (mm)}&80&35  \\
\text{Width / Outer Diameter (mm)}& 60 & $D$10  \\
\text{\space Height / Internal Diameter (mm)}\space& 50 & $D$6  \\
\text{Material}& N35 & N30  \\
\hline
%\multicolumn{4}{l}{$^{\mathrm{a}}$Sample of a Table footnote.}
\end{tabularx}
\label{tab2}
\end{center}
\end{table}

\subsection{Experimental Verification}
As shown in \eqref{eq7}, the friction model given in this paper can adapt to the relative motion of medium contact with different sliding friction coefficients. According to the friction coefficient of PVC and PLA materials \cite{ref17}\cite{ref18} and the tensile force of 0.042$\sim$0.048N, we set the sliding friction coefficient of PVC transparent plate and capsule robot as 0.22. According to our friction dynamic model, the difference of sliding friction coefficient does not affect the validity of the model.

\begin{figure*}[htbp]
\centerline{\includegraphics[width=7in]{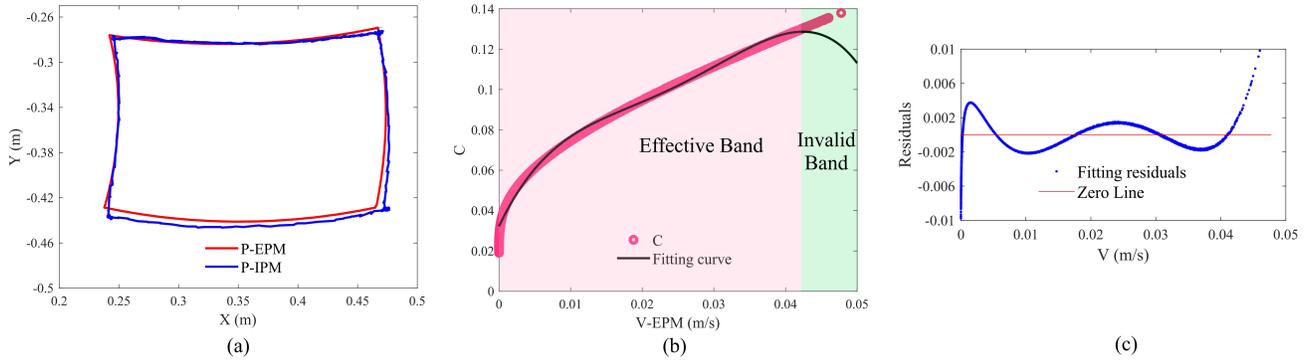}}
\caption{(a) The red line represents the EPM preset trajectory, and the blue line represents the trajectory of the IPM following movement. (b) The light red area indicates the boundary and the mixed fitting state, and the light green area indicates the state where the two curves are completely separated. (c) Residual diagram of ${\bf{v}}$-$c$ fitting.}
\label{fig7}
\end{figure*}

To ensure the rationality of the parameters in the given friction model, the coordinates of the end of the manipulator and the capsule robot are read and saved at the same frequency. The red line in Fig.~\ref{fig7}(a) indicates the preset trajectory of the manipulator. The IPM follows the EPM movement, which track point information is collected by the camera. The coordinate data (${{\bf{P}}_{\bf{E}}} \in {\mathbb{R}^3}$) of the EPM at the end of the robotic arm is directly read from the ros node. At the same time, the coordinate information of IPM and EPM is saved in the same program based on time series. We obtain the ${\bf{a}}$ (capsule acceleration) by obtaining the second-order derivative of time for the trajectory point (${{\bf{P}}_{\bf{I}}} \in {\mathbb{R}^3}$) information of the capsule robot. ${\bf{v}}$ is obtained by finding the first derivative of time for ${{\bf{P}}_{\bf{E}}}$. By substituting the distance of each (${{\bf{P}}_{\bf{E}}}$-${{\bf{P}}_{\bf{I}}}$) into the Maxwell simulation, ${\bf{F}}$ and  ${{\bf{F}}_{\bf{z}}}$ at the corresponding time can be obtained.

In order to learn the friction model in \eqref{eq7}, we repeat the above process to obtain 24786 track point data (including {\bf{F}}, ${{\bf{F}}_{\bf{z}}}$, ${\bf{a}}$). The trend of the discrete point graph is curved. Consequently, we use polynomial in Matlab Curve Fitting Toolbox \cite{MATLAB:2017b} to fit (as shown in Fig.~\ref{fig7}(b)). Besides, the residual diagram (Fig.~\ref{fig7}(c)) shows that the sum variance of ${\bf{v}}$-$c$ is 0.03977 when the control speed of the manipulator is between 0.005m/s and 0.04m/s. In this experimental environment, the specific fitting relationship is obtained as follows:
\begin{equation}
\begin{array}{l}
c = {p_1}*{{\bf{v}}^4} + {p_2}*{{\bf{v}}^3} + {p_3}*{{\bf{v}}^2} + {p_4}*{\bf{v}} + {p_5}\\
\left\{ \begin{array}{l}
{p_1} =  - 1.189e + 05\\
{p_2} = 1.173e + 04\\
{p_3} =  - 404.7\\
{p_4} = 7.44\\
{p_5} = 0.03222
\end{array} \right.
\end{array}\label{eq12}
\end{equation}

\subsection{Experimental Results}

\begin{figure*}[htbp]
\centerline{\includegraphics[width=7in]{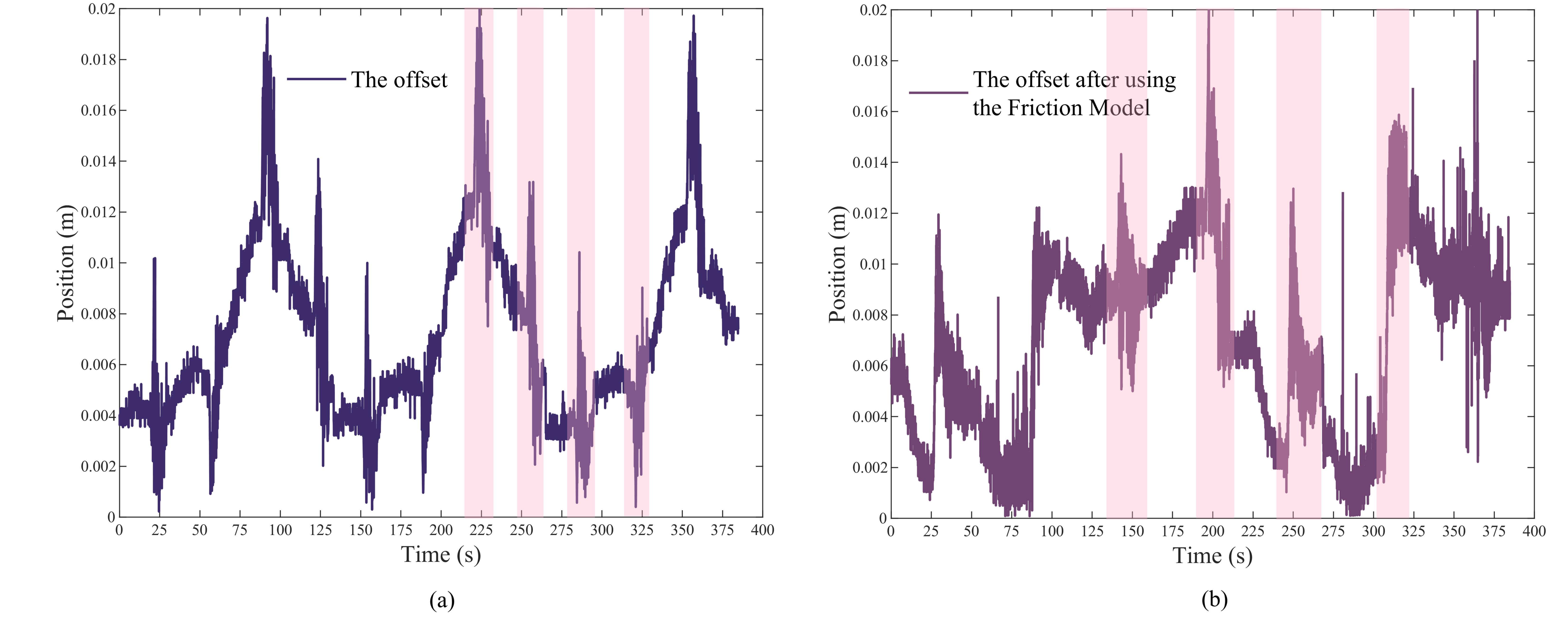}}
\caption{The shaded areas in (a)(b) indicate the process of movement. (a) shows the control error without considering the friction of the tether model. (b) represents the control error when considering the friction of the tether model.}
\label{fig8}
\end{figure*}

From \eqref{eq7} and \eqref{eq8} that the specific friction model of this experimental system is established. We analyze the trajectory’s STD (standard deviation) of the movement trajectory of the capsule robot in the shaded area in the Fig.~\ref{fig8}. As shown in Fig.~\ref{fig8}(a) that the  STD of the capsule robot is 3.5mm. Furthermore, we use the friction model to reduce 20\% speed increment of the manipulator and control the manipulator to complete the same preset trajectory in Fig.~\ref{fig7}. Similarly, we analyze the STD of the robot. It can be seen from Fig.~\ref{fig8}(b) that the STD dropped to 3.3mm after reduce the value of ${\bf{v}}$ in \eqref{eq6} to 0.8${\bf{v}}$. As shown in the shaded area in the Fig.~\ref{fig8}, we use the Learning friction model to reduce the STD by 5.6\% in a controlled experiment by calculating the absolute position error between the EPM and IPM in the process of motion.

\section{Discussion}
\begin{figure*}[htbp]
\centerline{\includegraphics[width=7in]{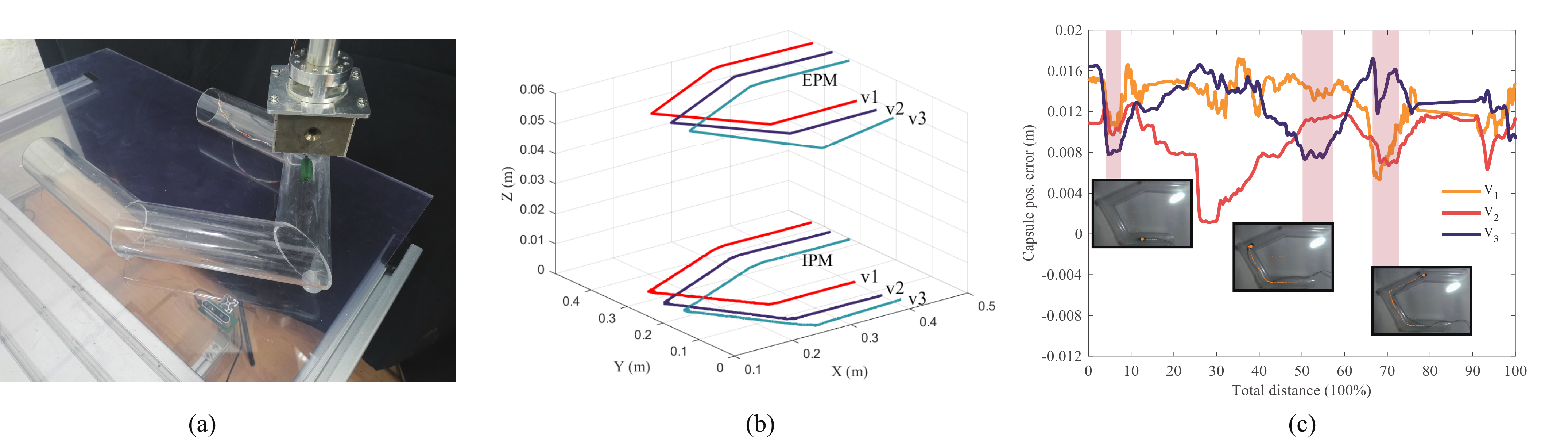}}
\caption{(a) Panorama of the experiment in the pipeline. (b) The trajectory diagram of the 10 averages of the 3 sets of speeds. The same color is a group of EPM and IPM trajectory control group, v1, v2, v3 are 0.045m/s, 0.02m/s, 0.005m/s respectively. (c) shows different errors at three speeds. The three shaded areas from left to right correspond to the starting segment, the second corner, and the third corner.}
\label{fig9}
\end{figure*}

The friction model mentioned in Letter has been validated in the plane. Subsequently, we added movement experiments in the pipe. The pipe material shown in the Fig.~\ref{fig9}(a) is the same as the PVC pipe mentioned above, and the friction coefficient between the pipe and the capsule robot is 0.22. Considering that velocity is the variable controlled by our model, the conditions (EPM, IPM, H) of the experimental scene remain constant. Three control experiments are carried out as shown in the Fig.~\ref{fig9}(b)(3 trajectory diagrams), where the velocities of v1 and v3 are beyond the velocity range of the friction model mentioned in the Letter, and STDS are 4.9mm and 5.1mm respectively. Fig.~\ref{fig9}(c) shows the error of the capsule robot with the v2, in the speed range of the friction modelwhich in the range of the frction and its STD is 4.7mm. 

%\begin{figure}[htbp]
%\centerline{\includegraphics[width=7in]{10}}
%\caption{The picture shows the trajectory diagram of the 10 averages of the 3 sets of speeds. The same color is a group of EPM and IPM trajectory control group, v1, v2, v3 are 0.045m/s, 0.02m/s, 0.05m/s respectively. }
%\label{fig10}
%\end{figure}

\section{Conclusion}
In this study, a learned friction model for tethered capsule robot is proposed. We design an experimental system, and the experimental results demonstrate  the effectiveness of the model. In addition, we can get the specific relationship between the control speed ${\bf{v}}$ and the friction factor $c$ in the process of the capsule robot moving in any plane contact medium. In the future, we will further study the space attitude control of the tethered capsule robots and optimize the structural components of the capsule robot to obtain better control accuracy.

\section*{ACKNOWLEDGMENTS}
This work was supported by the Natural Science Foundation of Jiangsu Province (Grant No. BK20180235).

%\caption{Authors' background}
\begin{table*}[htbp]
\section*{Authors' background}
\renewcommand\arraystretch{1.5}%设置行距
%\captionsetup{labelformat=empty}
\begin{center}
\begin{tabular}{|c|c|c|c|}
\hline
\textbf{Your Name}&\textbf{Title}&\textbf{Research Field}&\textbf{Personal website} \\
\hline
Miao Li\textsuperscript{*} & associate professor& robotics, machine earning and applied nonlinear control&\href{https://miaoli.github.io/}{https://miaoli.github.io/} \\
\hline
Yi Wang& master student & capsule robot &\href{http://aric.whu.edu.cn/team-member/master-2/wangyi/}{http://aric.whu.edu.cn/team-member/master-2/wangyi/} \\
\hline
\end{tabular}
\end{center}
\end{table*}

\bibliographystyle{IEEEtran}
\bibliography{reference}

\end{document}